\documentclass[
]{ceurart}

\usepackage{lstsemantic}
\usepackage[T1]{fontenc}
\usepackage[scaled=.9]{beramono}
\newtheorem{definition}{Definition}
\newtheorem{proposition}[definition]{Proposition}

\begin{document}

\copyrightyear{2022}
\copyrightclause{Copyright for this paper by its authors.
  Use permitted under Creative Commons License Attribution 4.0
  International (CC BY 4.0).}

\conference{NeSy 2022, 16th International Workshop on Neural-Symbolic Learning and Reasoning, Cumberland Lodge, Windsor, UK}

\title{Modular design patterns for neural-symbolic integration: refinement and combination}

\author[1]{Till Mossakowski}[%
orcid=0000-0002-8938-5204,
email=till.mossakowski@ovgu.de,
url=https://iks.cs.ovgu.de/~till,
]
\address[1]{Otto-von-Guericke-Universität Magdeburg, Germany}

\begin{abstract}
We formalise some aspects of the neural-symbol design patterns of van
Bekkum et al., such that we can formally define notions of
\emph{refinement} of patterns, as well as modular \emph{combination}
of larger patterns from smaller building blocks. These formal
notions have been implemented in the heterogeneous tool set (Hets),
such that patterns and refinements can be checked for well-formedness,
and combinations can be computed.
\end{abstract}

\begin{keywords}
  design pattern \sep
  refinement \sep
  combination \sep
  ontology
\end{keywords}

\maketitle

\section{Introduction}

The integration of subsymbolic and symbolic methods in artificial
intelligence, known as neural-symbolic integration, has been studied
for three decades now
\cite{DBLP:journals/ml/TowellS93,DBLP:books/daglib/0007534,DBLP:series/sci/2007-77,DBLP:series/cogtech/GarcezLG2009},
with a recently growing interest
\cite{HitzlerSarker22,DBLP:journals/ai/BadreddineGSS22,DBLP:journals/ml/ChenHJHAH21,DBLP:journals/corr/abs-2006-13155}.
While a solid theoretical basis is mostly lacking
\cite{vanHarmelen2022}, hundreds of specific methods and architectures
have been engineered and proven to be superior to both purely symbolic and to
purely subsymbolic methods. A certain structure has been brought into
this plethora of methods by developing classification schemas
\cite{VonRueden.ea:2021,DBLP:conf/ijcai/RaedtDMM20,DBLP:journals/apin/BekkumBHMT21}

We here follow neural-symbolic \emph{design patterns} developed in
\cite{DBLP:journals/apin/BekkumBHMT21}. They allow the description of
neural-symbolic systems using small building blocks like instances,
models, processes and actors in a modular way, and provide a useful
visualisation of the architecture of such systems. However, these design
patterns remain informal in \cite{DBLP:journals/apin/BekkumBHMT21}.
The goal of the present work is to formalise some aspects
of these design patterns, such that we can define notions of
\emph{refinement} of patterns, as well as modular \emph{combination}
of larger patterns from smaller building blocks.

Potential target audiences of our formalisation and tool support are
designers of neural-symbolic systems, researchers who want to create
post-hoc descriptions of such systems, as well as researchers who want to
relate and systemantically arrange such systems in some kind of
taxonomy.

\section{Neural-symbolic design patterns}

\begin{figure}
\hspace{-1cm}  \includegraphics[width=\textwidth]{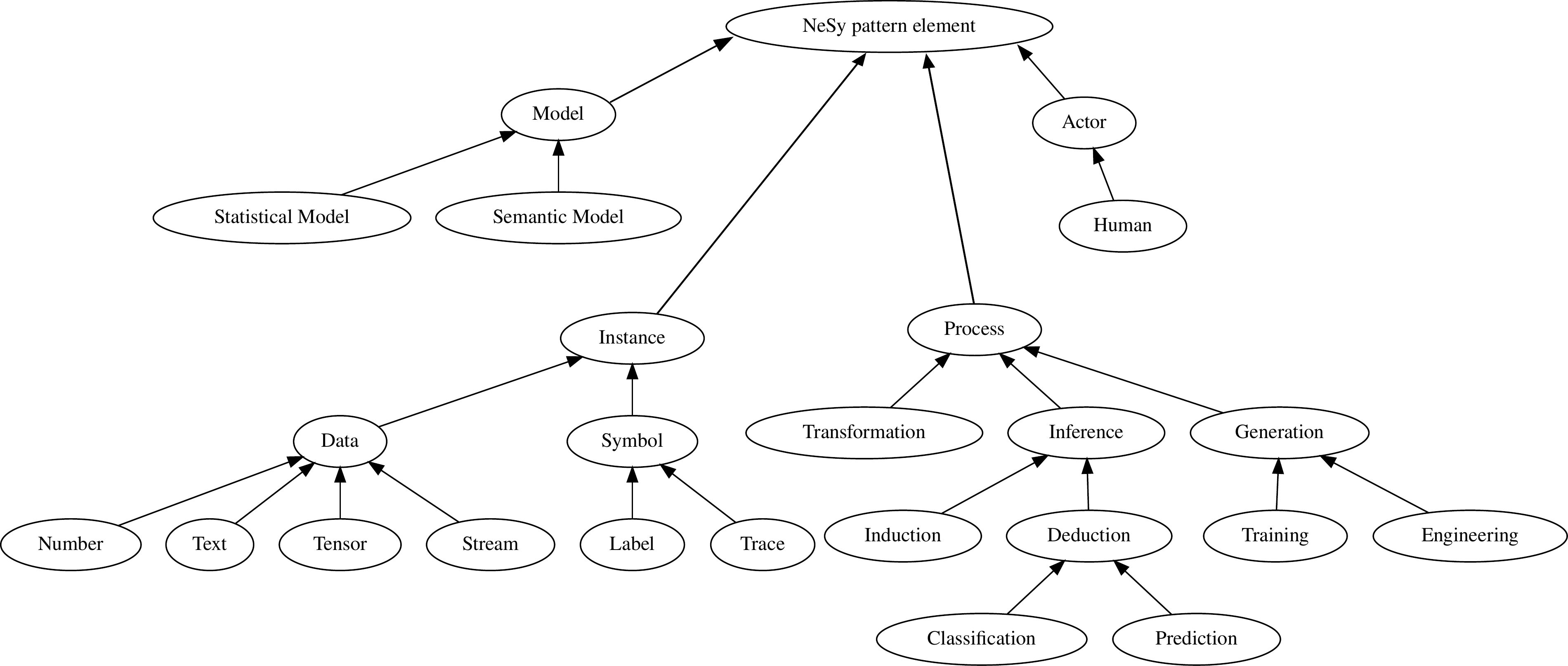}
  \caption{Ontology of pattern elements }
 \label{fig:onto}
\end{figure}

\begin{figure}
  \begin{tabular}{l}
  \includegraphics[scale=0.5]{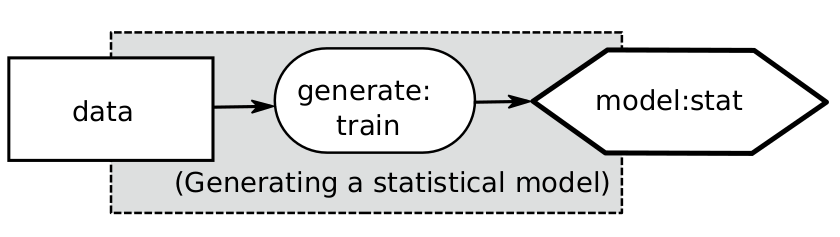}\hfill\\
  \includegraphics[scale=0.6]{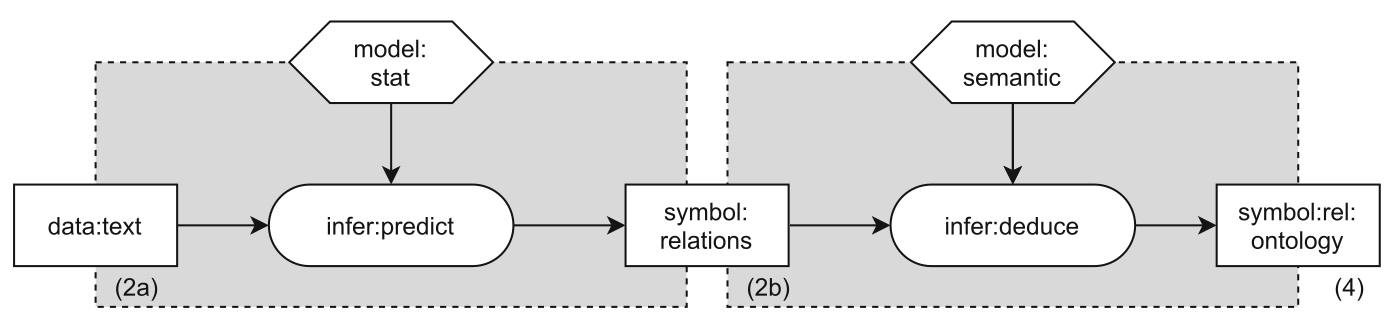}\hfill~
  \end{tabular}
  \caption{Two sample patterns} \label{fig:sample-patterns}
\end{figure}

The language for design patterns describing neural-symbolic systems
developed in \cite{DBLP:journals/apin/BekkumBHMT21} is based on a taxonomy
of pattern elements that has been introduced in the appendix of
\cite{DBLP:journals/apin/BekkumBHMT21}, and that is shown
in Fig.~\ref{fig:onto}. An extended version with more focus
on actors is presented in
\cite{DBLP:journals/corr/abs-2109-09331}. Note that the
top class in \cite{DBLP:journals/corr/abs-2109-09331}
is \texttt{Boxology Taxonomy}, which we here replace by \texttt{NeSy pattern element},
which more appropriately characterises the involved objects.
The classes just below the top class \texttt{NeSy pattern element} are the following:
\begin{description}
\item[Instance] Instances are symbols or data that comprise the input or output
  of neural symbol systems.\footnote{Ontologically speaking,
    these are not instances but elements of design patterns
    specifying that at this place of the system architecture,
    in the real system there will be an input or output of instances.}
\item[Model] Models can be hand-crafted or trained from instances
  (the latter is learning by induction). Existing models can be applied
  to instances (deduction).
\item[Process] Possible processes are training and deduction.
\item[Actor] Actors can be e.g.\ human beings that handcraft a model; \cite{DBLP:journals/corr/abs-2109-09331} provides a more fine-grained discussion of actors.
\end{description}
Fig.~\ref{fig:sample-patterns} shows a simple pattern that generates a
statistical model (like a neural network) from data, as well as a more
complex truly neural-symbolic pattern integrating a statistical and a
semantic model.  Both patterns stem from
\cite{DBLP:journals/apin/BekkumBHMT21}.

\section{Formalisation}

We now formalise the pattern language of
\cite{DBLP:journals/apin/BekkumBHMT21}. Patterns are some form of directed
graphs. At a first glance, it seems that these graphs are bipartite,
such that only edges between process nodes and non-process nodes are
possible. However, examples \cite{DBLP:journals/corr/abs-2109-09331}
show that this would be overly restrictive. Hence, we arrive at:

\begin{definition}\label{def:pattern}
Fix an ontology (class hierarchy) of pattern elements.  A
\emph{neural-symbolic design pattern} (NeSy pattern) is a simple directed graph,
where nodes $n$ are labeled with classes $lab(n)$ from the ontology.
\end{definition}

While this definition is dependent on an ontology of pattern elements,
it is of course desirable to achieve a standardisation here.  We have
formalised the ontology defined in the appendix of
\cite{DBLP:journals/apin/BekkumBHMT21} as an OWL2 ontology in
Manchester syntax at \url{https://ontohub.org/meta/NeSyPatterns.omn},
such that it can be used in NeSy patterns. However, for specific patterns,
one might feel the need to extend this ontology with new pattern
elements. Therefore, Def.~\ref{def:pattern} does not fix the ontology,
but rather allows any ontology. In practice, it will be desirable
to collect such ad-hoc extensions of the ontology and integrate
them into a standardised ontology, whenever it seems appropriate.


\medskip
During a development of NeSy patterns, one starts with rather
abstract patterns, which can later be refined towards a specific systems.
We formalise this using the notion of \emph{refinement}
of NeSy patterns.
Refinements map patterns using a graph homomorphism.
Labels can become more specific, i.e.\ move downwards in
the class hierarchy of the ontology.

\begin{definition} \label{def:refinement}
Given NeSy patterns $P_1$ and $P_2$ over the same ontology, a
\emph{refinement} $\varphi:P_1\to P_2$ is a homomorphism of unlabled
graphs $\varphi:P_1\to P_2$, such that for each node $n\in P_1$,
$lab(\varphi(n)) \leq lab(n)$ holds in the class hierarchy of the ontology.
\end{definition}

\begin{figure}
  \includegraphics[scale=0.5]{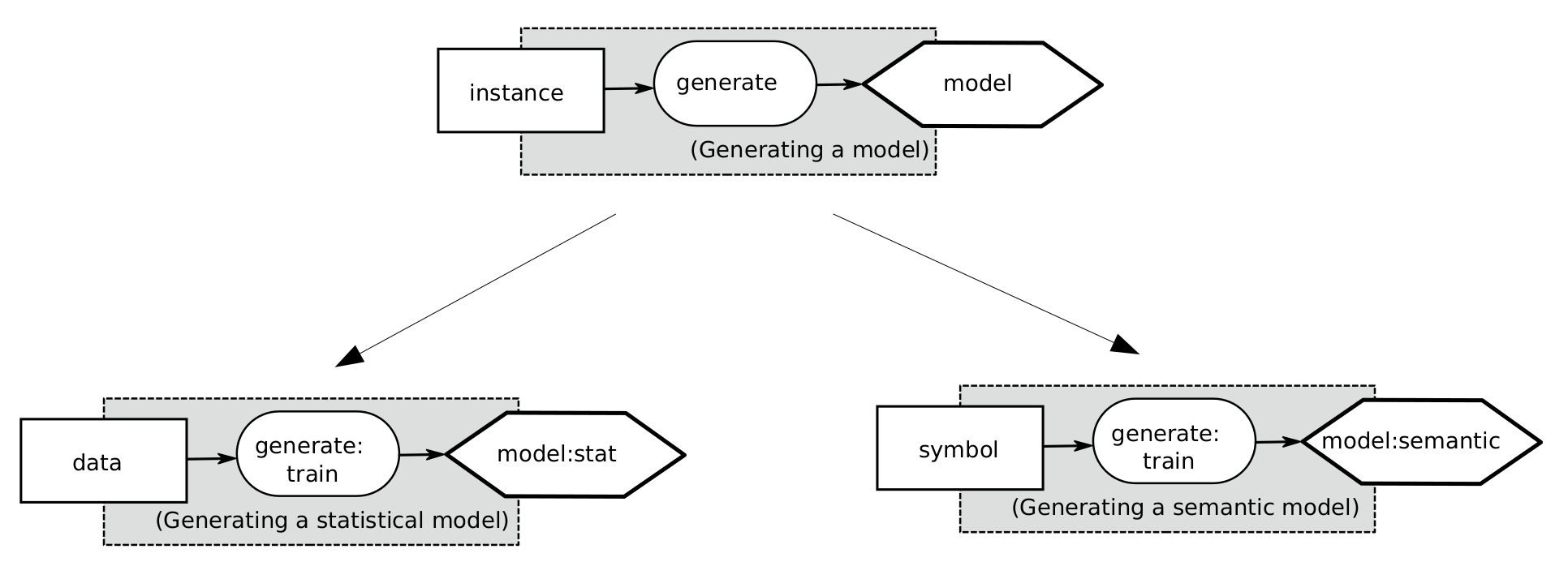}
  \caption{Two refinements of patterns} \label{fig:refinement}
\end{figure}

Fig.\ref{fig:refinement} shows an abstract pattern about generating
a model from instances, which is then refined in two ways:
first, into a pattern where a statistical model is generated
from data, and second, into a pattern where a semantic model is generated
from symbols.

\begin{figure}
  \includegraphics[scale=0.5]{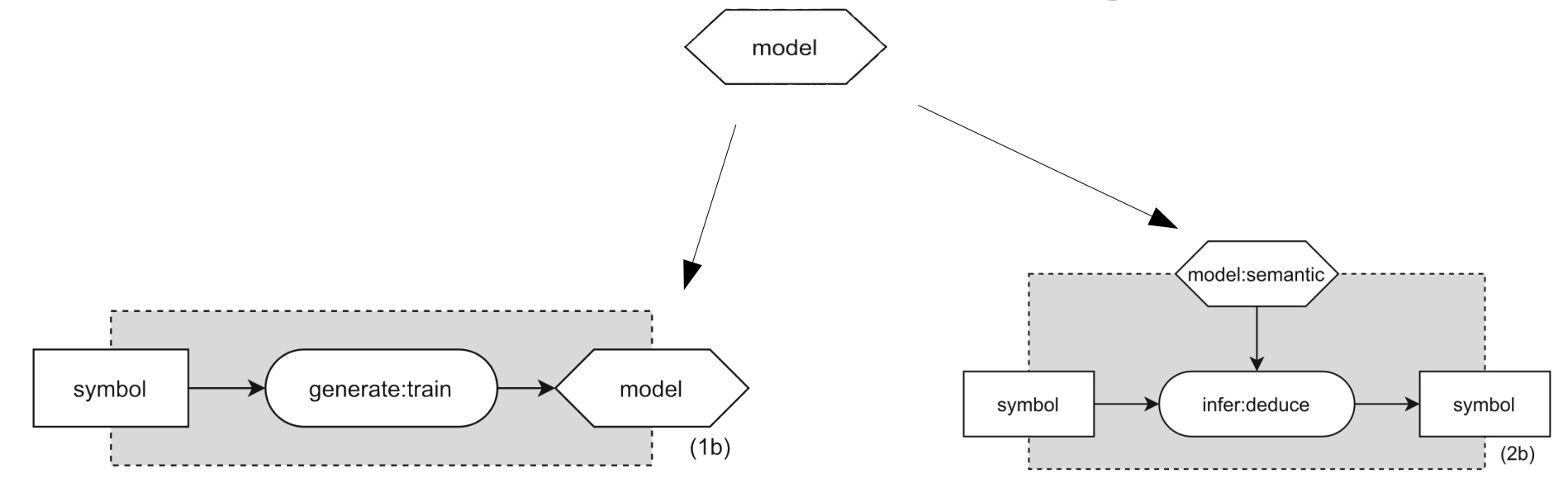}
  \caption{A network of three patterns and two refinements} \label{fig:network}
\end{figure}

\medskip
Patterns and refinements can be combined into \emph{networks}:
\vspace{-1ex}
\begin{definition}
  A \emph{network} consists of a graph with NeSy patterns as nodes and
  refinements as edges, showing how the patterns are interlinked.
  Type-correctness must hold, that is, an edge between pattern $P_1$
  and pattern $P_2$ must be a refinement from $P_1$ to $P_2$.
\end{definition}
Fig.~\ref{fig:network} shows such a network. The graph homomorphisms
of the refinement map the single model element of the upper pattern
to the model elements of the left and the right pattern, resp.
The importance of networks is that each network specifies a specific
way to glue together the patterns of the network into a combined
pattern.

Combination of patterns can be formally defined very succinctly
as \emph{colimits} in the sense of category theory \cite{AHS}:
\begin{definition}\label{def:combination}
Given a network of NeSy patterns and refinements, its
\emph{combination} (if existing) is the colimit in the category of
NeSy patterns and refinements.
\end{definition}

\begin{figure}
  \includegraphics[scale=0.5]{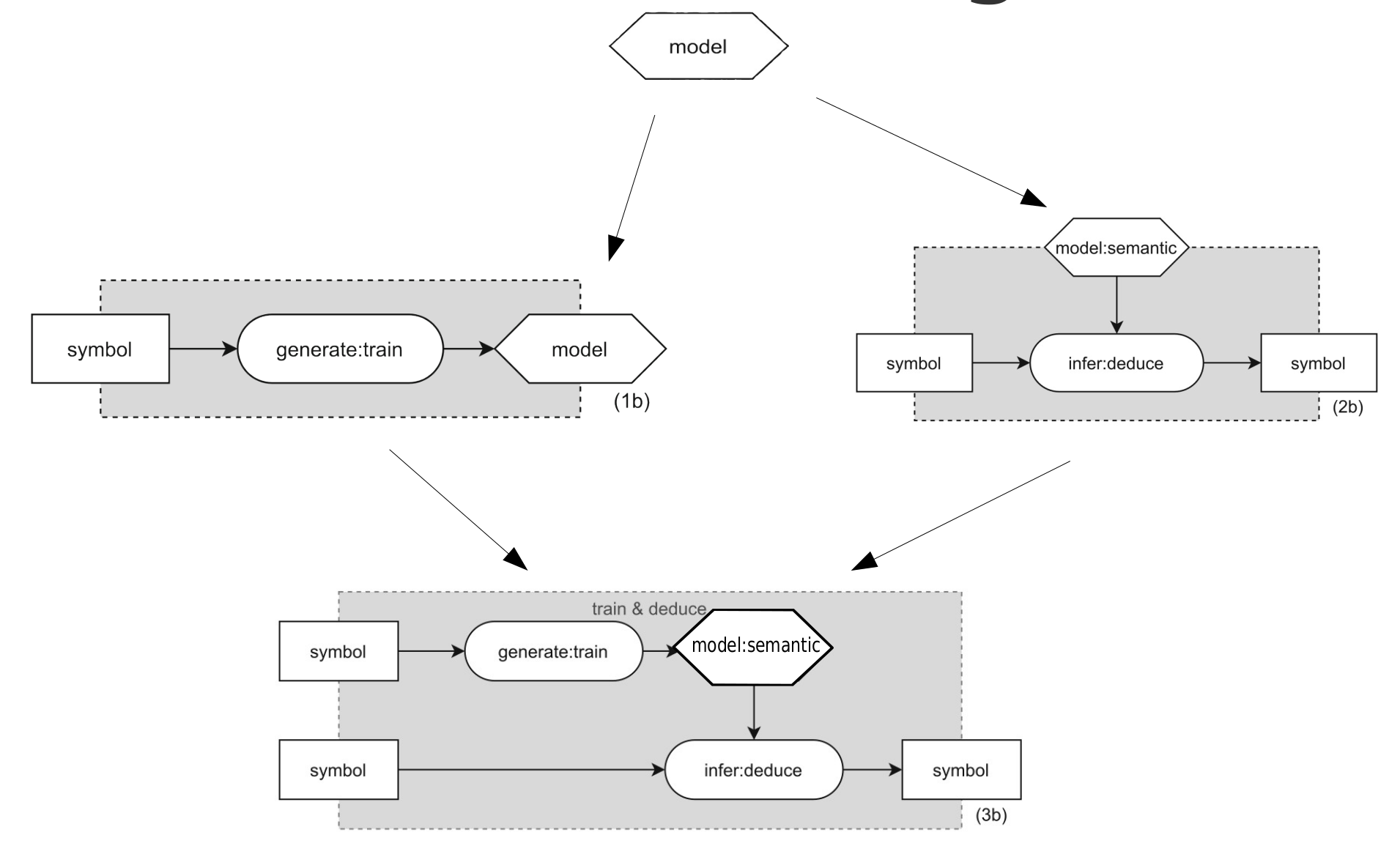}
  \caption{Combination of the network of Fig.~\ref{fig:network}.} \label{fig:combination}
\end{figure}

For readers not familiar with category theory, we will explain
this construction in more detail below. But let us first
look at some example. Fig.~\ref{fig:combination} shows a combination of two patterns. The
left pattern about generating a model is combined with the right
pattern about inferencing using a semantic model. The two patterns
are glued together at their model parts, using a very simple pattern
at the top, consisting just of a \texttt{Model}. The combination (shown at
the bottom) glues together the two input patterns. Note that the
model is a semantic model here, because the infimum of \texttt{Model}
and \texttt{Semantic model} in the ontology is \texttt{Semantic model}
\footnote{In Fig.~\ref{fig:combination}, annotations \texttt{model}
  and \texttt{model:semantic} are used instead of the proper ontology
  terms. In the future, this should be unified, either by providing
  additional class labels in the ontology, or by adapting the
  figures.}.

\bigskip
Now we give the explanation of the colimit construction in
Def.~\ref{def:combination} in elementary terms. The construction
is basically a glueing of all patterns of the network
at hand, realised by a disjoint union of node sets, which is
then quotiented as specified by the refinements in the network:

\begin{proposition}
The colimit as specified in Def.~\ref{def:combination} can be built as
follows. Given a network, let $(P_i)_{i\in I}$ be the family of patterns
of the network, and let $(\varphi_k)_{k\in K}$ be the family of refinements
of the network (each coming with a source and target pattern).
Let $N(P_i)$ be the set of nodes of pattern $P_i$.
Let
$$N = \biguplus_{i\in I}N(P_i)$$ be the disjoint union of all node
sets, and $\nu_i:N_i\to N$ (for $i\in I$) be the injections into the
disjoint union. Let $\sim$ be the equivalence relation over $N$ generated by
$$\nu_i(n) \sim \nu_j(\varphi_k(n))\ (n\in N_i)$$
for all $\varphi_k:P_i\to P_j$ in the network. Let
$$q:N\to N/{\sim}$$
be the factorisation of $N$ by $\sim$.

The colimit pattern $P$ has node set $N(P)=N/{\sim}$.
Let $\mu_i:N(P_i)\to N(P)$, defined
as $\mu_i= q \circ \nu_i$, be the injection of the $i$-th pattern
into the colimt. These injections together form the so-called colimit injections.
Edges in $P$ are all pairs $$(\mu_i(n_1),\mu_i(n_2))$$ for
$(n_1,n_2)$ an edge in pattern $P_i$. That is, $P$ contains
exactly those edges that are required to turn the colimit injections
into graph homomorphisms.

Finally, for a node $n \in N(P)$, let $N_n = \bigcup_{i\in
  I}\mu_i^{-1}(n)$.  Then
$$lab(n) = \inf_{m\in N_n} lab(m)$$

Now the colimit exists if the above infimum
$\inf_{m\in N_n} lab(m)$ exists for all nodes $n \in N(P)$ --- otherwise, the colimit is not defined.
\qed
\end{proposition}
Undefinedness of the colimit can arise for example if a node annotated
with \texttt{Model} is refined to (a) a node annotated with
\texttt{Semantic model} and (b) a node annotated with
\texttt{Statistical model}. Since there is no common subclass of
\texttt{Semantic model} and \texttt{Statistical model} in the
ontology, the above infimum does not exist. Note that this situation
can change if we change the ontology, i.e.\ by adding a term
\texttt{Hybrid model} as a subclass of both \texttt{Semantic
  model} and \texttt{Statistical model}. Indeed, such a term will
be very useful for specifying a pattern for logical neural networks
\cite{DBLP:journals/corr/abs-2006-13155}.

\section{Implementation}

The formal notions of patterns and refinements have been implemented
in the heterogeneous tool set (Hets)
\cite{DBLP:conf/tacas/MossakowskiML07}.\footnote{Hets is freely
  available under a GPL licence at
  \url{https://github.com/spechub/Hets}} Using the structuring meta
lanaguage DOL \cite{mossakowski2015distributed}\footnote{See also \url{https://dol-omg.org}}, patterns and
refinements can be written down as shown in Fig.~\ref{fig:dol}.  Using
the declaration \texttt{data ontohub:NeSyPatterns.omn},\footnote{The use
  of the \textbf{data} keyword is not related to the term
  \texttt{Data} of the ontology. It has historical reasons, because it
  is also used for linking process logics with logics for data in
  Hets.} each pattern refers to some ontology of pattern
elements. Here, \texttt{ontohub:NeSyPatterns} is a CURIE that
abbreviates the URL \url{https://ontohub.org/meta/NeSyPatterns} (using
a prefix declaration). Instead, the ontology could also be specified directly by a URL,
or an OWL2 ontology can be specified inline.

In the patterns, terms of the ontology have to be used exactly as they
are, while in the visualised patterns, often abbreviations and/or
annotations with superclasses are used. Also note that optional node
identifiers can be prepended with a colon (this similar to the
notation a:C in OWL2 ABoxes). This is important for distinguishing
nodes that are annotated with the same pattern element, and for
referencing nodes that have been declared earlier. E.g.\ the notation
\texttt{d : Deduction} in the example in Fig.~\ref{fig:dol} ensures
that there is only one node of type \texttt{Deduction}, and not
two.  The example also shows how the DOL language allows the
definition of networks of patterns and refinements. Such networks can
then be combined into a new pattern, in the sense of
Def.~\ref{def:combination}.

Hets will check these patterns, networks and combinations for
well-formedness, and the result of the combination can be
computed.\footnote{For details, see
  \url{https://github.com/spechub/Hets/wiki/Patterns-for-neural-symbolic-systems}}
For the example from Fig.~\ref{fig:dol}, a Hets development graph 
showing the different patterns (as nodes) and refinements (as edges)
is shown in
Fig.~\ref{fig:hets-example}. By clicking on the individual nodes,
one can display the individual NeSy patterns.
The upper arrow in Fig.~\ref{fig:hets-example} is a double
arrow. It is not a NeSy pattern refinement, but it rather
embeds the OWL2 ontology into an intermediate NeSy pattern node,
which is then used by the three declared NeSy patterns.

The modular design and re-use of NeSy patterns has been advocated in
\cite{DBLP:journals/apin/BekkumBHMT21}. With our formalisation in DOL,
we now can write down modular NeSy patterns in a precise syntax.  We
also have formalised all examples of
\cite{DBLP:journals/apin/BekkumBHMT21} in a library\footnote{See
  \url{https://github.com/spechub/Hets-lib/tree/master/NeSy}}, such
that system architects can re-use these patterns and build new ones on
top of them easily. The implementation in Hets allows us to flatten
complex modular designs and look at the resulting NeSy patterns.
Fig.~\ref{fig:dol2} shows one such pattern (pattern (2d) in \cite{DBLP:journals/apin/BekkumBHMT21}); it requires an extension of the OWL2 ontology,
because the class \emph{Embedding} has not been included in
the ontology so far.

\begin{figure}
  \begin{lstlisting}
%prefix( ontohub: <https://ontohub.org/meta/> )%
logic NeSyPatterns    
pattern Model = data ontohub:NeSyPatterns.omn
  Model;
end
pattern Train = data ontohub:NeSyPatterns.omn
  Symbol -> Training -> Model;
end
pattern SemanticDeduction = data ontohub:NeSyPatterns.omn
  Symbol -> d : Deduction -> Symbol;
  Semantic_Model -> d : Deduction; 
end
refinement R1 = Model refined to Train    
end
refinement R2 = Model refined to SemanticDeduction    
end
network N =
  Train, SemanticDeduction, R1, R2
end
pattern SemanticGenerateAndTrain =
  combine N
end
\end{lstlisting}
  \caption{The combination of Fig.~\ref{fig:combination}, formally specified in DOL} \label{fig:dol}
\end{figure}

\begin{figure}
  \centering
  \includegraphics[scale=0.35]{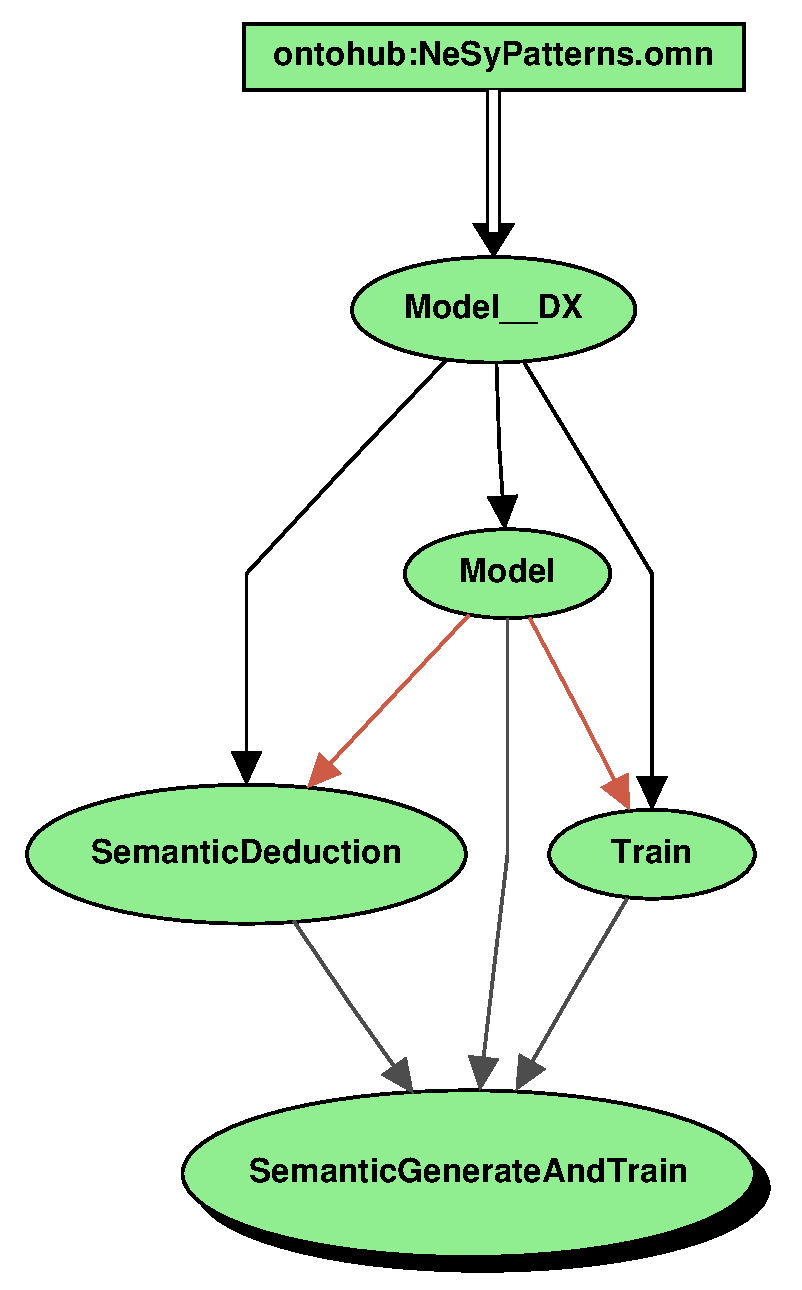}
  \caption{Hets development graph for the exmaple of Fig.~\ref{fig:dol}.} \label{fig:hets-example}
\end{figure}

\begin{figure}
  \begin{lstlisting}
%prefix( ontohub: <https://ontohub.org/meta/> )%
logic NeSyPatterns    
pattern Embedding =
  data { ontohub:NeSyPatterns.omn
         then Class Embedding SubClassOf: Transformation }  
  Symbol -> e:Embedding -> Data;  Semantic_Model -> e:Embedding;
end
\end{lstlisting}
  \caption{Pattern (2d) of \cite{DBLP:journals/apin/BekkumBHMT21}, formally specified in DOL} \label{fig:dol2}
\end{figure}
  
\section{Conclusion and future work}

We have formalised neural-symbolic design patterns using simple graphs
over some ontology of neural-symbolic elements.  We deliberately have
not used OWL2 ABoxes or RDF for formalising design patterns. Compared
to such a formulation, our formulation as simple graphs is simpler,
clearer and more concise to write. This also means that the suitable
notions of refinements and colimit are simpler than those for
OWL2 ABoxes or RDF.  Reasoning about refinement currently is done by
Hets' static
analysis, which checks the inequality of
Def.~\ref{def:refinement}. That said, we will provide a translation of
our pattern language into OWL2 ABoxes.  We will use the relations
(object properties) \texttt{providesInput} and \texttt{hasOutput} to
express edges in the graphs that comprise design patterns.  Our
notation \texttt{a : Symbol -> b : Training -> c : Model} would be
translated into
\begin{lstlisting}
a : Symbol
providesInput(a,b)
b : Training
hasOutput(b,c)
c : Model
\end{lstlisting}
which despite the use of concise description logic notation is still
far more verbose (and OWL2 syntax would be even more verbose).\footnote{The idea to formalise everything below
  \texttt{Process} as relation (object property), such that the above
  situation can be characterised by one triple b(a,c), is not
  ontologically valid, because processes are not relations. Moreover,
  in \cite{DBLP:journals/corr/abs-2109-09331}, links between
  proccesses are used, which could not be easily represented in this
  schema, while we can represent these, using a further relation like
  \texttt{throughput}.}

The formalisation of neural-symbolic patterns paves the way for
several useful developments. First, the use of a formal syntax for
patterns and of a formal ontology for pattern elements leads to
precision and standardisation. It would be useful to develop the
ontology of pattern elements as a joint community effort, such that it
can be referenced in patterns. Note that the DOL language allows the
local extension of the ontology, which (as we expect) will often be
needed. Moreover, DOL also allows the parallel refinement of a pattern
and its ontology, such that patterns written over different ontologies
can be refined and combined, too.  Of course, for the sake of
unification, such local ontology extensions could and should be later
integrated into the ontology, if found useful by the community.

Secondly, the ontology could also be used to impose constraints on
patterns. For example, using the above
sketched translation to an OWL2 ABox, we could state that
only machine learning models can be trained by adding the axiom
$$ \exists \text{hasOutput}^{-1}. \text{Training} \sqsubseteq
\text{'Statistiscal Model'} $$ This axiom can be used to reason about
patterns, with the outcome that e.g.\ \texttt{Statistiscal Model} need
to be refined into \texttt{Model}.  A further axiom stating
disjointness of \texttt{Statistiscal Model} and \texttt{Semantic
  Model}\footnote{However, note that such an axiom is debatable,
  because there are hybrid models like Logical neural networks
  \cite{DBLP:journals/corr/abs-2006-13155} that could be considered to
  be both statistical and semantic.} would make patterns featuring a
training of a semantic model inconsistent, which can be found be OWL2
reasoners. Hence, axioms can help to ensure the internal consistency
of patterns.

As noted in \cite{DBLP:journals/apin/BekkumBHMT21}, patterns could be
also used to specific and build real neural-symbolic systems in a
modular way. A step towards this goal would be to equip pattern
elements with signatures. For symbols, this would be a first-order
signature, such that symbols could be represented as terms over that
signature. For data tensors, one would specify their dimensions.  Such
signatures could then be combined into more complex signatures for the
specification of models and processes. Once such signatures have been
provided, the next step will be to use logical languages like the
Hoare-style logic of \cite{DBLP:conf/ijcai/XieKN22} for the
axiomatic specification and verification of neural-symbolic systems.

\paragraph{Acknowledgements} The author wants to thank Fabian Neuhaus
for useful feedback and discussions, the anonymouos
reviewers for useful suggestions and Mihai Codescu and Björn Gehrke
for working on the Hets implementation.

\bibliography{nesy}

\end{document}